\documentclass[10pt,twocolumn,letterpaper]{article}

\usepackage{iccv}              

\usepackage{multirow}
\usepackage{amsmath}
\usepackage{color}
\usepackage{pifont}

\usepackage{makecell}

\usepackage{adjustbox}
\usepackage{graphicx}
\usepackage{fix-cm}
\makeatletter
\def\thickhline{%
  \noalign{\ifnum0=`}\fi\hrule \@height \thickarrayrulewidth \futurelet
   \reserved@a\@xthickhline}
\def\@xthickhline{\ifx\reserved@a\thickhline
               \vskip\doublerulesep
               \vskip-\thickarrayrulewidth
             \fi
      \ifnum0=`{\fi}}
\makeatother
\newlength{\thickarrayrulewidth}
\usepackage{array}
\newcolumntype{L}[1]{>{\raggedright\let\newline\\\arraybackslash\hspace{0pt}}m{#1}}
\newcolumntype{C}[1]{>{\centering\let\newline\\\arraybackslash\hspace{0pt}}m{#1}}
\newcolumntype{R}[1]{>{\raggedleft\let\newline\\\arraybackslash\hspace{0pt}}m{#1}}

\definecolor{iccvblue}{rgb}{0.21,0.49,0.74}
\usepackage[pagebackref,breaklinks,colorlinks,allcolors=iccvblue]{hyperref}

\def\paperID{5133} 
\def\confName{ICCV}
\def\confYear{2025}

\title{Robust Adverse Weather Removal via Spectral-based Spatial Grouping}

\author{Yuhwan Jeong$^{*}$, Yunseo Yang$^{*}$, Youngho Yoon$^{*}$, and Kuk-Jin Yoon  \\
Visual Intelligence Lab., KAIST, Korea\\
{\tt\small \{jeongyh98,acorn,dudgh1732,kjyoon\}@kaist.ac.kr}
}
\begin{document}
\maketitle
\def\thefootnote{*}\footnotetext{Denotes equal contribution. \\ The code is available at \url{github.com/jeongyh98/SSGformer}.}\def\thefootnote{\arabic{footnote}}

\begin{abstract}

Adverse weather conditions cause diverse and complex degradation patterns, driving the development of All-in-One (AiO) models.
However, recent AiO solutions still struggle to capture diverse degradations, since global filtering methods like direct operations on the frequency domain fail to handle highly variable and localized distortions.
To address these issue, we propose Spectral-based Spatial Grouping Transformer (SSGformer), a novel approach that leverages spectral decomposition and group-wise attention for multi-weather image restoration. 
SSGformer decomposes images into high-frequency edge features using conventional edge detection and low-frequency information via Singular Value Decomposition.
We utilize multi-head linear attention to effectively model the relationship between these features.
The fused features are integrated with the input to generate a grouping-mask that clusters regions based on the spatial similarity and image texture. 
To fully leverage this mask, we introduce a group-wise attention mechanism, enabling robust adverse weather removal and ensuring consistent performance across diverse weather conditions.
We also propose a Spatial Grouping Transformer Block that uses both channel attention and spatial attention, effectively balancing feature-wise relationships and spatial dependencies.
Extensive experiments show the superiority of our approach, validating its effectiveness in handling the varied and intricate adverse weather degradations. 

\end{abstract}  
\vspace{-5pt}
\section{Introduction}
\label{sec:intro}

\begin{figure}[t!]
    \centering
    \includegraphics[width=0.995\linewidth]{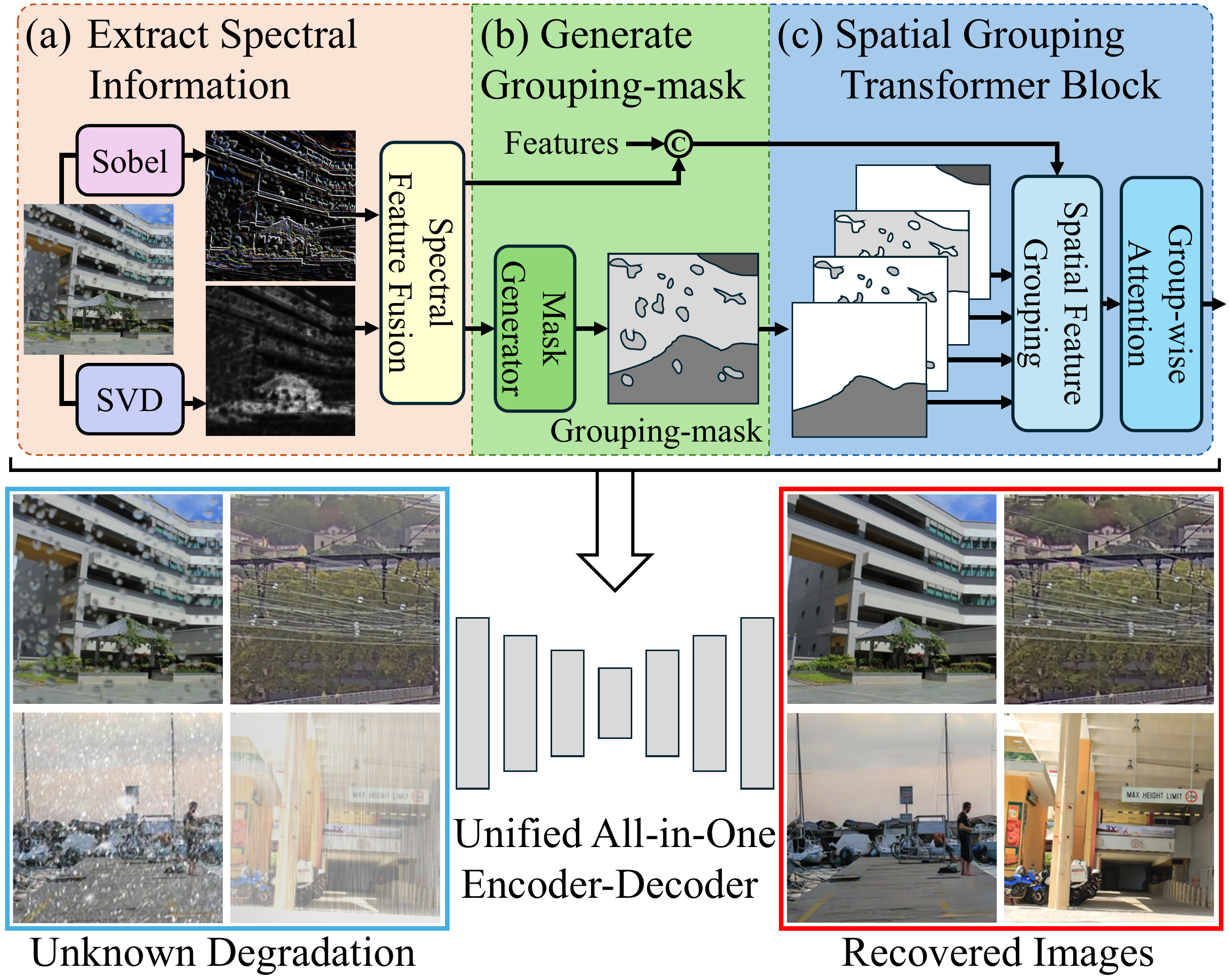}
    \vspace{-17pt}
    \caption{\textbf{Overview.} We propose SSGformer, an All-in-One adverse weather removal model with three key components. (a) Sobel operator and SVD extract spectral information from degraded image, producing degradation-aware features through spectral feature fusion. (b) The fused features identify degradation patterns and generate a ``grouping-mask" for spatial grouping. (c) Features are grouped by the mask for group-wise spatial/channel attention.}
    \vspace{-16pt}
    \label{fig:teaser}
\end{figure}

Adverse weather conditions, such as snow, rain, raindrop, and haze, severely degrade visual information, making image restoration an essential task for various applications~\cite{gupta2024robust, yang2024semantic, lee2022learning}. 
To address this issue, researchers have explored methods to remove weather-induced degradation from images~\cite{li2019heavy,zhang2019image,yang2019single,yasarla2019uncertainty,qu2019enhanced,liu2018desnownet,chen2020jstasr,qian2018attentive,quan2019deep}. 
In particular, significant efforts have been made to leverage image-based priors, such as edges and fourier patterns, as a foundation for restoring degraded images~\cite{yeh2018rain, Zhu_2017_ICCV, zhu2015fast, kang2011automatic, schechner2001instant}.

Early research on adverse weather removal primarily focused on handling single weather conditions~\cite{chen2023learning, yang2022self, song2023vision, chen2023msp}. However, as the demand for comprehensive models to address multiple types of degradation grew, the focus shifted toward All-in-One (AiO) approaches for adverse weather removal~\cite{potlapalli2024promptir,zhang2023ingredient,park2023all,cui2023selective,lin2024improving,li2022all,valanarasu2022transweather,chen2022learning,zhu2023learning,ozdenizci2023restoring,ye2023adverse,patil2023multi,yang2024genuine,zhang2024efficient,zhu2024mwformer}.

Despite these advancements, developing a unified model capable of handling diverse weather conditions remains a major challenge due to the highly variable nature of weather-induced degradations. The complex interactions between different atmospheric conditions lead to unpredictable visual degradations, making it difficult to create a single framework that generalizes well across all scenarios.

To address the challenges posed by adverse weather conditions, several studies have explored image restoration techniques in the frequency domain. AIRFormer~\cite{gao2023frequency}, a general-purpose AiO model, constructs a frequency-guided transformer encoder by incorporating wavelet-based prior information to enhance feature extraction, leveraging structured frequency priors to improve restoration quality.
Fourmer~\cite{zhou2023fourmer}, similarly, employs the Fourier transform to disentangle image degradation from content components, utilizing the global nature of the Fourier domain to process degradation patterns comprehensively.
AdaIR~\cite{cui2024adair} is designed to identify and mitigate degradation patterns through frequency domain analysis.
These methods show the potential of frequency-based restoration, which transforms an image into the frequency domain and manipulates its frequency components.
However, operations in the frequency domain that affect the entire region modify the frequency content of the entire image.
This global filtering method may be suitable for addressing repetitive patterns, such as blur or noise. However, it is not effective for handling random and localized degradations that occur in adverse weather conditions.
Additionally, while the wavelet transform processes images by dividing them into frequency bands, it reduces the image resolution by half. 
Consequently, to use this frequency information, the image needs to be resized back to its original dimensions, which may introduce distortions.
Therefore, an effective solution is required that extracts valuable spectral information for frequency-based image restoration, while preserving spatial details to prevent the loss of crucial contextual information.

To conquer this problem, we propose SSGformer, which not only extracts essential features using image-based priors but also ensures their effective integration within a comprehensive framework.
As shown in~\cref{fig:teaser} (a), the first step extracts edge and low-rank information from the input image while preserving spatial details.
Conventional edge detection captures high-frequency edge features, while Singular Value Decomposition (SVD) analyzes degradation textures in the low-frequency domain.
To model the relationships between spectral information, we use multi-head linear attention to capture the interdependencies within the spectral features.
Building on this, we expand the information exchange beyond local spectral regions by identifying and comparing spatially similar areas across the image, facilitating broader degradation-aware feature sharing and more efficient grouping.
To achieve the grouping, we employ a mask generator to capture spatial similarity from the spectral information and generate a grouping-mask that clusters relevant features based on their values (\cref{fig:teaser} (b)).

To effectively utilize these grouped features and facilitate global information exchange, we design a specialized architecture tailored to optimize their integration.
In this architecture, we propose attention mechanisms that operate on mask-based grouped features exhibiting similar spatial characteristics as shown in~\cref{fig:teaser} (c).
These group-wise attention is applied both spatially and across channels, allowing the model to focus on the most relevant features.
Consequently, our proposed architecture enables robust degradation removal across diverse adverse weather conditions.

Our contributions can be summarized as:
\begin{itemize}
  \item We propose SSGformer, an innovative transformer architecture that enhances restoration effectively handling adverse weather conditions. 
  \item We reveal that directly extracting image information via edge detection and SVD is effective for robust adverse weather removal. Moreover, efficient grouping of this information is key to performance gains.
  \item Our extensive experiments showcase the robust restoration capability of SSGformer, achieving state-of-the-art performance across a complex weather degradation.
\end{itemize}


\section{Related works}
\label{sec:relworks}

\subsection{Task-specific adverse weather removal}
Attempts to restore images degraded by adverse weather conditions to clean images initially began as single-task approaches. Common restoration tasks involves deraining, desnowing, dehazing, and raindrop removal.

\noindent
\textbf{Deraining} aims to remove rain streaks from images. 
In the early stages, the task was solved using hand-crafted or rule-based methods~\cite{kang2011automatic, chen2013generalized}. The development of single image deraining has entered the deep learning era, leading to the emergence of numerous deep learning-based approaches~\cite{fu2017removing, fu2023continual, chen2023hybrid, wang2023smartassign}. Some of works utilized a conditional GAN~\cite{li2019heavy, zhang2019image}, edge loss~\cite{yang2019single}, and a multi-scale residual network~\cite{yasarla2019uncertainty} for deraining.
Since transformers emerged~\cite{vaswani2017attention}, studies~\cite{xiao2022image, chen2023learning} have leveraged ViT's image analysis capabilities~\cite{dosovitskiy2020image}.

\noindent
\textbf{Dehazing} has been widely studied across various fields, thereby aiding daily life. Early methods~\cite{cai2016dehazenet, ren2016single} utilized CNNs for dehazing. Proposed learning methods include considering both atmospheric light and transmission map~\cite{li2017aod}, GAN~\cite{qu2019enhanced, yang2022self}, and contrastive learning~\cite{Wu_2021_CVPR}.

\noindent
\textbf{Desnowing}~\cite{liu2018desnownet, quan2023image, chen2023snow,chen2023msp} tasks remove snow particles from an image and fill in the empty areas behind them. The research has been proposed using a method that is aware of size and transparency~\cite{chen2020jstasr}, semantic and geometric priors~\cite{zhang2021deep}, and dual-tree wavelet transform~\cite{chen2021all}.

\noindent
\textbf{Rain drop removal}
has been studied using temporal information from videos~\cite{you2015adherent} in the early era.
During the machine learning era, CNNs~\cite{eigen2013restoring} were initially used to clean images, followed by enhancements through GAN-based methods~\cite{qian2018attentive, yan2022raingan}. More research~\cite{guo2020joint, yang2020raindrop} has been widely proposed using edge information~\cite{quan2019deep}, multi-scale attention~\cite{shao2021uncertainty}, and transformer~\cite{song2023vision}.

\begin{figure*}[t]
    \centering
    \includegraphics[width=0.995\linewidth]{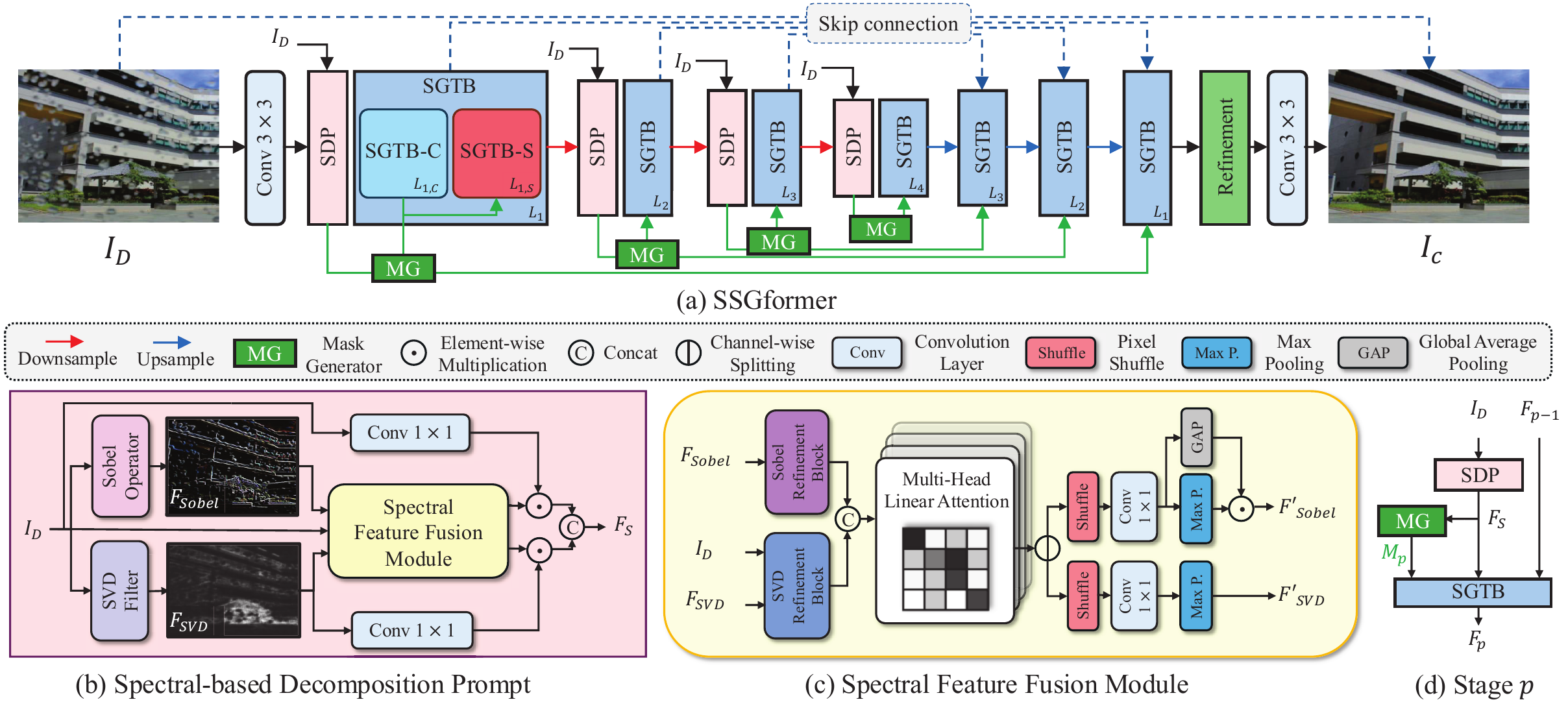}
    \vspace{-7pt}
    \caption{\textbf{Overall framework.} (a) SSGformer is a transformer-based encoder-decoder network that restores a weather-degraded image $I_D$ to a clean image $I_C$. The encoder consists of Spectral-based Decomposition Prompt (SDP), Mask Generator (MG), and Spatial Grouping Transformer Block (SGTB), which include two types of blocks: SGTB-C and SGTB-S. At stage $p$, there exist as many blocks as the number indicated by $L_p$. The stage $p$ increases from 1 to 4 and then decreases back to 1 in the decoder.
    (b) SDP takes the input image $I_D$, passes it through Sobel operator and SVD filter to generate $F_{Sobel}$ and $F_{SVD}$, which are processed by the Spectral Feature Fusion Module to produce $F_S$. (c) In the Spectral Feature Fusion Module, $F_{Sobel}$, $F_{SVD}$, and $I_D$, are passed through multi-head linear attention to obtain the fused features $F'_{Sobel}$ and $F'_{SVD}$. (d) Feature flow at $p$-th stage in the encoder. $F_S$ from SDP is sent to both MG and SGTB, while the previous stage output $F_{p-1}$ and the ``grouping-mask" $M_p$ from MG are fed into SGTB to produce $F_{p}$.}
    \vspace{-14pt}
    \label{fig:architecture}
\end{figure*}

\subsection{All-in-One model for adverse weather removal}
Early restoration methods were constrained by the need for separate, task-specific models for each adverse weather condition. This limitation has led to efforts to develop unified frameworks. Li~\emph{et al.}~\cite{li2020all} introduced task-specific encoders, which were later expanded upon with contrastive learning~\cite{li2022all}. Subsequent advancements include two-stage knowledge distillation~\cite{chen2022learning}, single encoder-decoder architectures~\cite{valanarasu2022transweather}, and the decoupling of general and specific weather features~\cite{zhu2023learning}.
Ye~\emph{et al.}~\cite{ye2023adverse} utilized quantized codebook prior to remove degradation, while Sun~\emph{et al.}~\cite{sun2025restoring} further improved restoration quality by grouping similar features using pixel-sorting.
However, in some challenging scenarios, intensity may not sufficiently distinguish semantically different regions, possibly affecting attention behavior.
Although recent methods~\cite{yang2024language,luo2023controlling} incorporate external knowledge through LLMs or VLMs, our approach is fundamentally grounded in intra-image information without resorting to any external source.

\section{Methods}
\label{sec:methods}


\subsection{Overview}

Our goal is to restore images damaged by random degradations and highly diverse weather conditions within a unified framework.
Many existing AiO models address severe weather degradation through global filtering but struggle with highly variable conditions.
To solve this object, one of our key idea is that adding comparisons between similar features through grouping can lead to more robust adverse weather removal.
Moreover, grouping features based on spectral characteristics provides valuable information for image restoration.
With these ideas, we propose a novel architecture, \textbf{SSGformer}, that captures inherent variability and randomness through spectral analysis and effectively integrates them within a comprehensive framework.

SSGformer is a transformer-based unified 4-stage encoder-decoder network designed to restore a weather-degraded image $I_D$ to a clean image $I_C$. 
The encoder consists of a Spectral-based Decomposition Prompt (SDP), a Mask Generator (MG), and a Spatial Grouping Transformer Block (SGTB). 
The decoder incorporates SGTB and a refinement block.
For more details, please refer to ~\cref{fig:architecture} (a).

To consider the pixel similarity, it is necessary to group the features, which requires a criterion for dividing the groups.
Therefore, we begin the design process selecting the information to determine the rule.
To extract information for the criterion, SDP is designed to perform spectral analysis using two types of filters, the Sobel operator and SVD filter, and fuse the features with multi-head linear attention.
Then, the MG generates a grouping-mask from degradation-aware feature $F_S$, the output of the SDP, enabling the spatial grouping of features with similar characteristics.
Using this mask, SGTB applies group-wise attention to enhance feature interactions.
Depending on the type of attention used, SGTB is categorized into SGTB-C for channel attention, and SGTB-S for spatial attention.
Detailed descriptions of SDP, MG, and SGTB are provided in \cref{sec:SDP}, \cref{sec:mask}, and \cref{sec:SGTB}, respectively.


\subsection{Spectral-based Decomposition Prompt}
\label{sec:SDP} 
Spectral-based Decomposition Prompt (SDP) (\cref{fig:architecture} (b)) is divided into two stages. In the first stage, spectral information of the input image $I_D$ is decomposed using multiple filters. Subsequently, we integrate these features through our Spectral Feature Fusion Module and convolution layers, yielding the degradation-aware feature $F_S$.

\noindent\textbf{Spectral analysis.}
To extract essential priors related to spectral information, we employ two decomposition filters.
One is the Sobel operator which detects high frequency information by highlighting intensity changes. The other, Singular Value Decomposition (SVD), finds low frequency knowledge of the degraded image.
The features obtained from the two filter operations, $F_{Sobel}$ and $F_{SVD}$, both in $\mathbb{R}^{H \times W \times 1}$, capture different information from the same input $I_D$. $H$ and $W$ denote the spatial dimensions at current stage.
This complementary spectral analysis enables early capture of various types of degradation, which can provide valuable spectral features for effective restoration.

\noindent\textbf{Spectral Feature Fusion Module.}
To find the relationship between each spectral feature, we adopt multi-head linear attention~\cite{katharopoulos2020transformers}, which improves feature representation by sharing spectral information with different characteristics.
Prior to this, we introduce the Sobel and SVD refinement blocks to enhance each spectral feature (see \cref{fig:architecture} (c)).
The Sobel refinement block uses a convolution layer and feature reorganization, which refine information by incorporating surrounding context.
The SVD refinement block consists of convolution layers, feature reorganization, and a deformable convolution layer, which refine the low-frequency components with $I_D$.
The refined features are then passed through multi-head linear attention, which is followed by channel splitting before further processing.
Afterward, we apply pixel shuffle~\cite{shi2016real} to upscale the feature, expanding the receptive field to better understand the context. The upscaled features are processed using max pooling or global average pooling to emphasize salient features and capture global context information, yielding $F'_{Sobel}$ and $F'_{SVD}$.

The manipulated spectral features are combined to form the degradation-aware feature $F_S$. Specially, $F'_{Sobel}$ is element-wise multiplied with the feature extracted from $I_{D}$, while the $F'_{SVD}$ is multiplied with the $F_{SVD}$ after a convolution layer. These two components are then integrated as:
\begin{equation}
F_S = [F'_{\text{Sobel}} \odot \text{Conv}_{1}(I_{D}),\: F'_{\text{SVD}} \odot \text{Conv}_{1}(F_{SVD})]
\end{equation}
where $\odot$ denotes element-wise multiplication. $[\bullet,\bullet]$ is the concatenation operation.

\subsection{Mask Generator}
\label{sec:mask}
Previous methods~\cite{cui2024adair, li2023prompt, sun2025restoring} typically rely on the feature derived from the prompt without any additional processing, often simply concatenating it and feeding it directly into the network. However, these approaches do not fully utilize the extracted information. To overcome this, we introduce a spatial grouping method that first generates a grouping-mask from the degradation-aware feature $F_S$ to enhance feature utilization. This mask identifies regions in the feature map with similar characteristics, enabling effective spatial grouping. It plays a crucial role in focusing on regions that share similar degradation patterns, thereby improving the efficiency and effectiveness of the restoration process.
At stage $p$, the mask generator consists of a simple convolution layer that synthesizes information across multiple channels to produce a single-channel mask, $M_p = \text{Conv}_{7}(F_S),~\text{where}~M_p\in \mathbb{R}^{H \times W \times 1}$.
Notably, the mask is used in both the encoder and decoder at the same stage, ensuring coherent and effective feature manipulation.

With this grouping-mask $M_p$, feature elements are divided into $g_p$ number of groups, structuring them for further processing. The mask utilization is described in~\cref{sec:FGA}. 

\begin{figure}[t]
    \centering
    \includegraphics[width=0.995\linewidth]{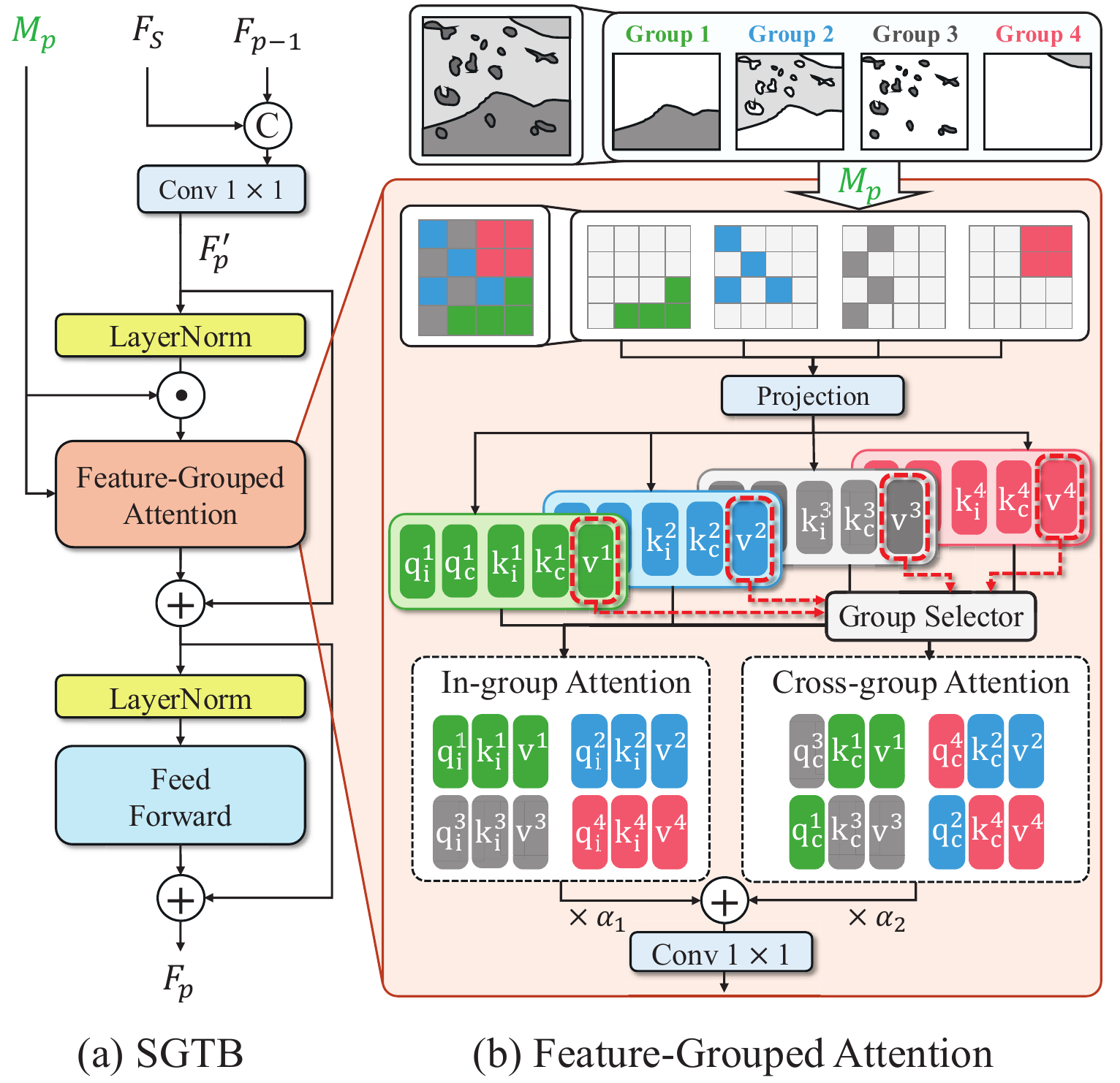}
    \vspace{-16pt}
    \caption{(a) \textbf{Spatial Grouping Transformer Block (SGTB)} starts with a convolution layer that fuses the previous stage features $F_{p-1}$ and the degradation-aware features $F_S$. The grouping-mask $M_p$ is element-wise multiplied with $F'_p$ and is also fed into Feature-Grouped Attention. (b) \textbf{Illustration of Feature-Grouped Attention.} Features are grouped using $M_p$, which encodes spatial similarity. In-group and cross-group attention are applied simultaneously, with a group selector handling query exchanges before cross-group attention. The same color represents the same group. This is an example when the number of groups, $g_p$ is 4.}
    \vspace{-7pt}
    \label{fig:SGTB}
\end{figure}

\subsection{Spatial Grouping Transformer Block}
\label{sec:SGTB}
One of our key idea is analyzing spatially correlated features and considering their contextual relationships can be helpful to handle the unpredictability of weather-induced distortions.
To this end, we design the Spatial Grouping Transformer Block (SGTB), which performs group-wise attention by leveraging spatial similarity information from generated masks.
Furthermore, unlike existing transformer-based AiO models that rely solely on channel attention, SGTB incorporates not only channel attention but also spatial attention, enhancing restoration performance by facilitating information exchange along the spatial axis.

SGTB follows a standard transformer structure, comprising a normalization layer, attention, and feed-forward network, as shown in \cref{fig:SGTB} (a).
At the beginning of each SGTB, the features $F_{p-1}$ from the previous stage (for the first block, the input degraded image) and the degradation-aware features $F_S$ extracted from SDP are combined through a convolution layer as follows:
\begin{equation}
F'_p= \text{Conv}_{1}[F_{p-1}, F_S].
\end{equation}
The resulting feature $F'_p$ goes through a normalization layer, Feature-Grouped Attention (FGA), and a feed-forward network (FFN), and is expressed as follows:
\vspace{-2pt}
\begin{align}
\hat{F}_{p} &= F'_p + \text{FGA}(\text{LayerNorm}(F'_p) \odot M_p, \ M_p), \\
F_{p} &= \hat{F}_{p} + \text{FFN}(\text{LayerNorm}(\hat{F}_{p})).
\end{align}

In FGA, we employ two attention mechanisms: channel (SGTB-C) and spatial attention (SGTB-S). The grouped features undergo attention operations, either channel-wise, by emphasizing the most relevant channels, or spatially, by focusing on key regions within the feature map. These attention operations are structurally identical, with the only distinction being the dimension of the attention matrix.

\begin{figure}[t]
    \centering
    \includegraphics[width=0.995\linewidth]{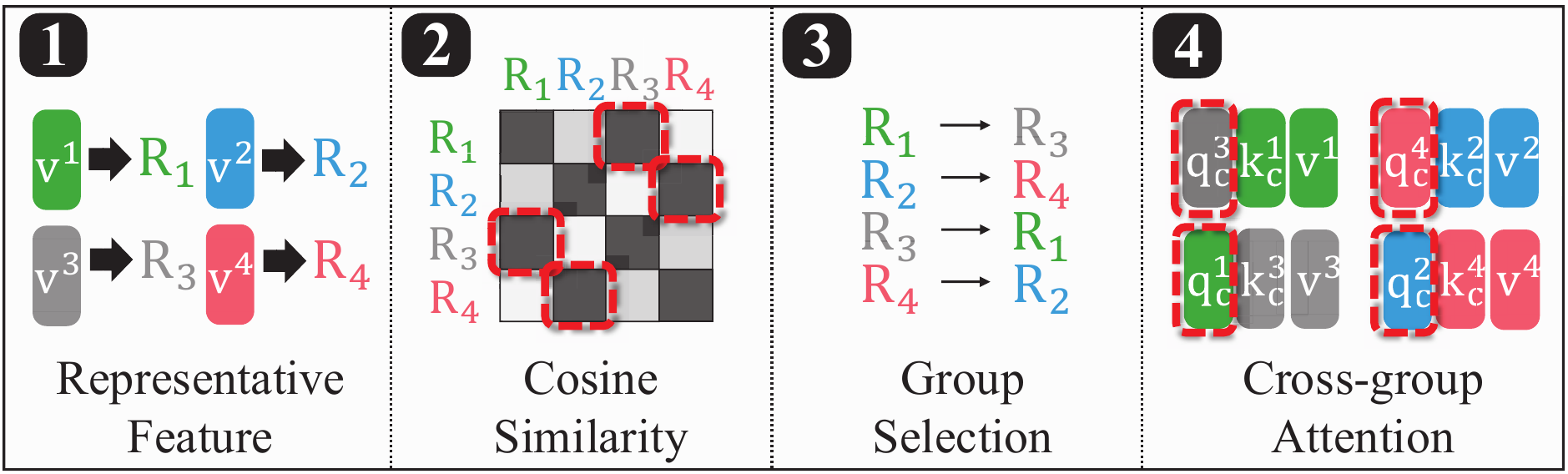}
    \vspace{-17pt}
    \caption{\textbf{Illustration of the group selector process.} (1) Each group's value is pooled to obtain a representative feature ($R_{1 \sim 4}$). (2) Cosine similarity is computed using the representative features to measure inter-group similarity.  The darker the matrix, the higher the similarity score. (3) The most similar group is selected. (4) Attention is computed using the query from the selected group. For clarity, the example is illustrated with four groups, where the same color represents the same group.}
    \vspace{-14pt}
    \label{fig:selector}
\end{figure}

\noindent
\textbf{Feature-Grouped Attention},
\label{sec:FGA}
which is the key component, utilizes the grouping-mask $M_p$ to perform spatial grouping of features.
We group features based on mask values, dividing them evenly into $g_p$ groups ($e.g.$, $g_p=4$ in \cref{fig:SGTB} (b)). Each grouped feature has the same spatial dimensions, $i.e.$, the same number of elements.
This grouping mechanism enables restoration by focusing on regions with similar characteristics, effectively handling spatial variability.

While attention within individual groups allows the model to capture local feature relationships, it is limited to learning only within those specific groups. To overcome this limitation, we extend the attention mechanism to capture cross-group relationships, enabling the model to learn broader feature dependencies and expand its receptive field.

Accordingly, we implement two types of group attention mechanisms, in-group and cross-group.
In-group attention is applied within each feature group, while cross-group attention identifies the relationship between itself and another group.
As shown in \cref{fig:SGTB} (b), to perform both types of attention simultaneously, the grouped features are projected to generate two queries, two keys, and one value:
\vspace{-2pt}
\begin{equation}
q_i^m, q_c^m, k_i^m, k_c^m, v^m = \text{Projection}(G_p^m),
\vspace{-2pt}
\end{equation}
where $G_p^m \in \mathbb{R}^{\frac{H \times W}{g_p} \times C}$ represents the feature of the $m$-th group at stage $p$, with $m \in \{1,2,...,g_p\}$.
The subscript $i$ indicates in-group attention, and $c$ indicates a feature used in cross-group.
The dimension of $G_p^m$ is derived by spatially grouping the original feature map, which has dimensions $\mathbb{R}^{H \times W \times C}$, where $C$ represents the channel dimension.
Each projected feature has the same dimensions as $G_p^m$.
Next, the in-group attention is calculated within the group as follows:
\begin{equation}
A_{\text{in}}^m = \text{In-Group}(q_i^m, k_i^m, v^m).
\end{equation}
For cross-group attention, rather than performing comparisons across all groups, which would be a computationally heavy task, we aim to efficiently target a specific group. 
Therefore, to identify the most relevant group, we propose the group selector which computes the similarity between all $v^m$ and selects the most relevant group index for group $m$ while excluding itself, as follows:
\begin{equation}
a_m = \text{GroupSelector}(\{{v^n}\}_{n=1}^{g_p}), \;\; s.t. \;\;  n \neq m 
\end{equation}
and $a_m$ denotes the selected group index (Details are described in the next section.)
The attention is performed using the query from the selected group, $q_c^{a_m}$:
\begin{equation}
A_{\text{cross}}^m = \text{Cross-Group}(q_c^{a_m}, k_c^m, v^m).
\end{equation}

After calculating $A_{\text{in}}^m$ and $A_{\text{cross}}^m$ for each group, they are aggregated to obtain the final attention outputs, $A_{\text{in}}$ and $A_{\text{cross}}$.
The output for FGA is combined with skip-connection as follows:
\begin{equation}
\hat{F}_{p} = F'_p + \text{Conv}_{1}(\alpha_1 \odot A_{\text{in}} + \alpha_2 \odot A_{\text{cross}}).
\end{equation}
Here, $\alpha_1$ and $\alpha_2$ are learnable weights that balance the contributions of $A_{\text{in}}$ and $A_{\text{cross}}$.

\begin{table*}[thb!]
\centering
\caption{Quantitative comparisons on the All-weather dataset~\cite{li2020all}. Methods with a ${}^{*}$ mark are adopted from All-in-One image restoration and trained with the All-weather dataset. The best values are marked in \textbf{bold}, while the second-best values are \underline{underlined}.}
\vspace{-7pt}
\resizebox{.995\textwidth}{!}{
\begin{tabular}{l|lcc|lcc|lcc|cc}
\thickhline
                            &         \multicolumn{3}{c|}{Rain}  &         \multicolumn{3}{c|}{Snow}           &         \multicolumn{3}{c|}{Raindrop}   &  \multicolumn{2}{c}{Average}        \\
                            & Method & PSNR & SSIM & Method & PSNR & SSIM & Method & PSNR & SSIM & PSNR & SSIM \\ \thickhline
\multirow{5}{*}{\rotatebox{90}{Single}} & CycleGAN~\cite{zhu2017unpaired} & 17.62 & 0.6560 & SPANet~\cite{wang2019spatial} & 23.70 & 0.7930 & pix2pix~\cite{isola2017image} & 28.02 & 0.8547 & - & -\\
                            & pix2pix~\cite{isola2017image} & 19.09 & 0.7100 & JSTASR~\cite{chen2020jstasr} & 25.32 & 0.8076 & RaindropAttn~\cite{quan2019deep} & 31.44 & 0.9263 & - & - \\
                            & HRGAN~\cite{li2019heavy} & 21.56 & 0.8550 & DesnowNet~\cite{liu2018desnownet} & 27.17 & 0.8983 & AttentiveGAN~\cite{qian2018attentive} & 31.59 & 0.9170 & - & - \\
                            & MPRNet~\cite{zamir2021multi} & 28.03 & 0.9192 & DDMSNet~\cite{zhang2021deep} & 28.85 & 0.8772 & MAXIM~\cite{tu2022maxim} & 31.87 & 0.9352 & - & - \\ 
                            & Fourmer~\cite{zhou2023fourmer} & 26.58 & 0.8740 & Fourmer~\cite{zhou2023fourmer} & 27.38 & 0.8555 & Fourmer~\cite{zhou2023fourmer} & 25.79 & 0.8500 & 26.58 & 0.8508 \\ \hline
\multirow{11}{*}{\rotatebox{90}{All-in-One}} & Restormer$^{*}$~\cite{zamir2022restormer} & 30.03 & 0.9215 & Restormer$^{*}$~\cite{zamir2022restormer} & 30.36 & 0.9068 & Restormer$^{*}$~\cite{zamir2022restormer} & 32.18 & 0.9408 & 30.86 & 0.923\\
                            & PromptIR$^{*}$~\cite{potlapalli2024promptir} & 30.49 & 0.9263 & PromptIR$^{*}$~\cite{potlapalli2024promptir} & 30.91 & 0.9148 & PromptIR$^{*}$~\cite{potlapalli2024promptir} & 32.56 & 0.9426 & 31.32 & 0.928\\
                            & AdaIR$^{*}$~\cite{cui2024adair} & 30.85 & 0.9286 & AdaIR$^{*}$~\cite{cui2024adair} & 31.01 & 0.9160 & AdaIR$^{*}$~\cite{cui2024adair} & 32.87 & 0.9432 & 31.58 & 0.929\\ 
                            & All-in-One~\cite{li2020all} & 24.71 & 0.8980 & All-in-One~\cite{li2020all} &  28.33 & 0.8820 & All-in-One~\cite{li2020all} & 31.12 & 0.9268 & 28.05 & 0.902 \\
                            & TransWeather~\cite{valanarasu2022transweather} & 28.83 & 0.9000 & TransWeather~\cite{valanarasu2022transweather} & 29.31 & 0.8879 & TransWeather~\cite{valanarasu2022transweather} & 30.17 & 0.9157 & 29.44 & 0.901 \\
                            & TUM~\cite{chen2022learning} & 29.27 & 0.9147 & TUM~\cite{chen2022learning} & 30.22 & 0.9071 & TUM~\cite{chen2022learning} & 31.81 & 0.9309 & 30.43 & 0.918 \\
                            & WGWS~\cite{zhu2023learning} & 29.32 & 0.9207 & WGWS~\cite{zhu2023learning} & 30.16 & 0.9007 & WGWS~\cite{zhu2023learning} & 32.38 & 0.9378 & 30.62 & 0.920 \\
                            & AWRCP~\cite{ye2023adverse} & 31.39 & 0.9329 & AWRCP~\cite{ye2023adverse} & 31.92 & \textbf{0.9341} & AWRCP~\cite{ye2023adverse} & 31.93 & 0.9314 &31.75 & 0.933 \\
                            & Histoformer~\cite{sun2025restoring} & \underline{32.08} & \underline{0.9389} & Histoformer~\cite{sun2025restoring} & \underline{32.16} & 0.9261 & Histoformer~\cite{sun2025restoring} & \underline{33.06} & \underline{0.9441} & \underline{32.43} & \underline{0.936} \\
                            & Ours & \textbf{32.43} & \textbf{0.9410} & Ours & \textbf{32.22} & \underline{0.9272} & Ours & \textbf{33.24} & \textbf{0.9489} & \textbf{32.63} & \textbf{0.939}\\ \thickhline
\end{tabular}}
\vspace{-14pt}
\label{tab:main}
\end{table*}

\noindent\textbf{Group selector.}
To identify the most relevant group, each group generates a representative feature $R_m$ by channel-wise pooling its values $v^m$, which aggregates the features across the channel dimension.
The cosine similarity is then computed between these representative features, and the most similar group is selected based on the highest similarity score.
This allows the model to adaptively establish cross-group relationships, even if the groups were not initially grouped together.
The process is illustrated step by step in ~\cref{fig:selector}.
The insight is that interacting with the most similar group allows the model to achieve more robust restoration by effectively leveraging relationships beyond the initial grouping.


\noindent\textbf{Attention configuration.}
Existing AiO models primarily employ channel attention, which performs attention mechanism along the channel dimension. 
While channel attention focuses on feature-wise relationships, it does not explicitly account for spatial dependencies.
In adverse weather conditions, spatial attention is crucial as it directly captures relationships in space, making it more effective for restoration tasks where degraded or occluded areas need to be reconstructed using surrounding context.
However, as image resolution increases, the computational cost of spatial attention grows quadratically.
To balance the benefits of spatial attention and computational efficiency, half of the SGTB blocks at each stage use channel attention (SGTB-C), while the other half use spatial attention (SGTB-S).
In each stage, SGTB-C is applied first and followed by SGTB-S.

\begin{table}[t]
    \centering
    \caption{Experiments on WeatherStream dataset. Trans. and Histo. denote TransWeather~\cite{valanarasu2022transweather} and Histoformer~\cite{sun2025restoring}, respectively.}
    \vspace{-7pt}
    \resizebox{.995\linewidth}{!}{
    \begin{tabular}{l|cc|cc|cc|cc}
    \thickhline
    & \multicolumn{2}{c|}{Rain} & \multicolumn{2}{c|}{Fog} & \multicolumn{2}{c|}{Snow} & \multicolumn{2}{c}{Avg.} \\
    Model  & PNSR & SSIM & PNSR & SSIM & PNSR & SSIM & PNSR & SSIM\\ \thickhline
    Trans.~\cite{valanarasu2022transweather} & 25.72 & 0.7911 & 24.30 & 0.6979 & 23.66 & 0.7386 & 24.56 & 0.742 \\
    WGWS~\cite{zhu2023learning}    & 25.75 & 0.7928 & 23.88 & 0.6752 & 23.66 & 0.7383 & 24.43 & 0.735 \\
    Histo.~\cite{sun2025restoring} & \underline{25.85} & \underline{0.8215} & \underline{24.36} & \textbf{0.7348}  & \underline{23.72} & \underline{0.7631} & \underline{24.64} & \underline{0.773} \\
    Ours  & \textbf{26.01} & \textbf{0.8218} & \textbf{24.39} & \underline{0.7335} & \textbf{23.91} & \textbf{0.7652} & \textbf{24.77} & \textbf{0.774} \\
    \thickhline
    \end{tabular}}
    \vspace{-14pt}
    \label{tab:main_ws}
\end{table}

\subsection{Objective function}
\label{sec:obj_func}

We train our model with the combination of $L1$ loss, $\mathcal{L}_{rec}$, and Pearson correlation loss~\cite{cohen2009pearson, sun2025restoring}, $\mathcal{L}_{cor}$ to consider pixel-level and patch-level correlations together.
The total objective function we used is as:
\vspace{-5pt}
\begin{equation}
    \mathcal{L} = \mathcal{L}_{rec} + \beta \mathcal{L}_{cor},
\end{equation}
where $\beta$ is the weight of the correlation loss. More about objective function are described in the supplementary.

\section{Experiments}
\label{sec:exp}
\subsection{Experimental settings}
\textbf{Datasets.}
We use two dataset to evaluate our model. One is a synthetic dataset, All-weather dataset~\cite{li2020all}, which contains Outdoor-rain~\cite{li2019heavy}, Snow100K-L~\cite{liu2018desnownet}, and Raindrop~\cite{qian2018attentive} dataset as previous researches~\cite{li2022all, valanarasu2022transweather, sun2025restoring} use. The other is a real dataset, WeatherStream~\cite{zhang2023weatherstream}, which contains the scenes of rain, snow, and fog. Each sub-dataset has A, B, and C degraded images. For fair comparisons, we follow the dataset utilization protocol of previous works~\cite{li2020all, sun2025restoring, zhu2023learning}.

\begin{figure*}[tbh]
    \centering
    \includegraphics[width=0.995\linewidth]{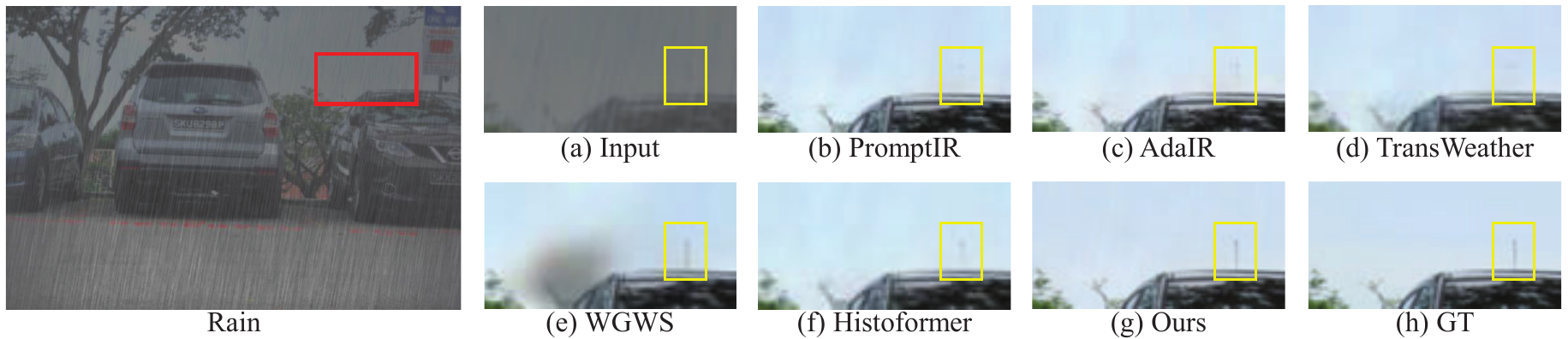}
    \vspace{-7pt}
    \caption{Qualitative results of deraining on Outdoor-rain~\cite{li2019heavy} dataset.}
    \vspace{-7pt}
    \label{fig:qual_rain}
\end{figure*}

\begin{figure*}[tbh]
    \centering
    \includegraphics[width=0.995\linewidth]{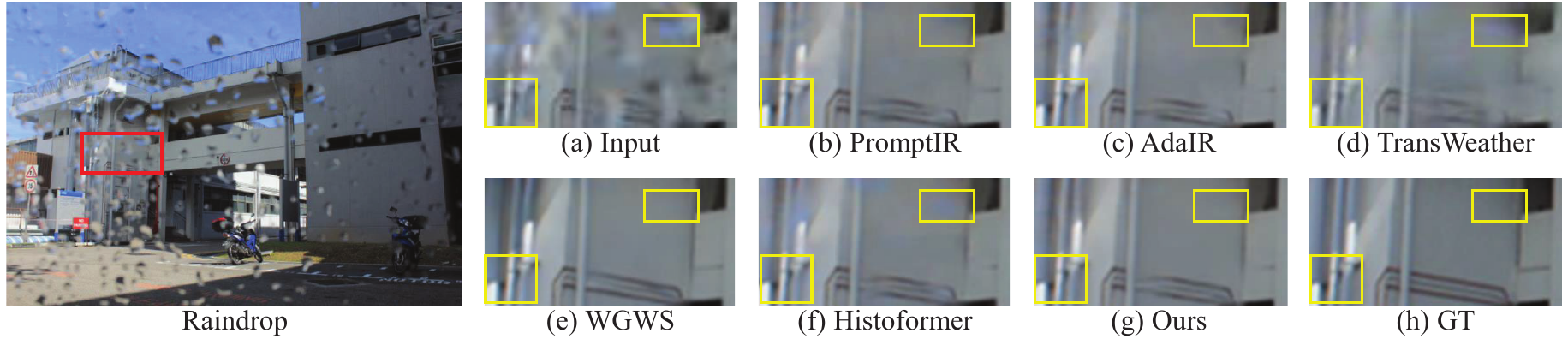}
    \vspace{-7pt}
    \caption{Qualitative results of raindrop removal on RainDrop~\cite{qian2018attentive} dataset.}
    \vspace{-14pt}
    \label{fig:qual_drop}
\end{figure*}

\noindent
\textbf{Baseline methods.}
To contextualize our results, we compare with single-task methods for raindrop removal (pix2pix~\cite{isola2017image}, RaindropAttn~\cite{quan2019deep}, AttentiveGAN~\cite{qian2018attentive}, MAXIM~\cite{tu2022maxim}, Fourmer~\cite{zhou2023fourmer}), snow removal (SPANet~\cite{wang2019spatial}, JSTASR~\cite{chen2020jstasr}, DesnowNet~\cite{liu2018desnownet}, DDMSNet~\cite{zhang2021deep}, Fourmer~\cite{zhou2023fourmer}), and rain removal (CycleGAN~\cite{zhu2017unpaired}, pix2pix~\cite{isola2017image}, HRGAN~\cite{li2019heavy}, MPRNet~\cite{zamir2021multi}, Fourmer~\cite{zhou2023fourmer}). We also evaluate against AiO restoration models (Restormer~\cite{zamir2022restormer}, PromptIR~\cite{potlapalli2024promptir}, AdaIR~\cite{cui2024adair}) and AiO adverse weather removal models (All-in-One~\cite{li2020all}, TransWeather~\cite{valanarasu2022transweather}, TUM~\cite{chen2022learning}, WGWS~\cite{zhu2023learning}, AWRCP\cite{ye2023adverse}, Histoformer~\cite{sun2025restoring}) that use only images as input without any external knowledge, such as language models, trained with previous strategies~\cite{li2020all,sun2025restoring} for All-weather dataset~\cite{li2020all}. Experiments on WeatherStream dataset~\cite{zhang2023weatherstream}, we compare our methods with TransWeather, WGWS, and Histoformer, which are recent AiO models.

\subsection{Comparisons with the state-of-the-arts}
\noindent
\textbf{Quantitative evaluation.}
In~\cref{tab:main}, we quantitatively compare our method with other single weather removal models, general AiO models, and AiO adverse weather removal models on the All-Weather dataset.
Our method achieves the best performance on rain and raindrop, outperforming the second-best method by +0.35dB and +0.18dB in PSNR, respectively, and the highest PSNR with the second-best SSIM in snow.
On average, it demonstrates the best overall performance on the All-Weather dataset.

Moreover,~\cref{tab:main_ws} presents the comparison on the WeatherStream dataset~\cite{zhang2023weatherstream}. Our SSGformer consistently achieved top performance across most cases, ranking second only in SSIM for fog degradation. Our method performs the best on average in the real-world dataset.

\begin{figure*}[tbh]
    \centering
    \includegraphics[width=0.995\linewidth]{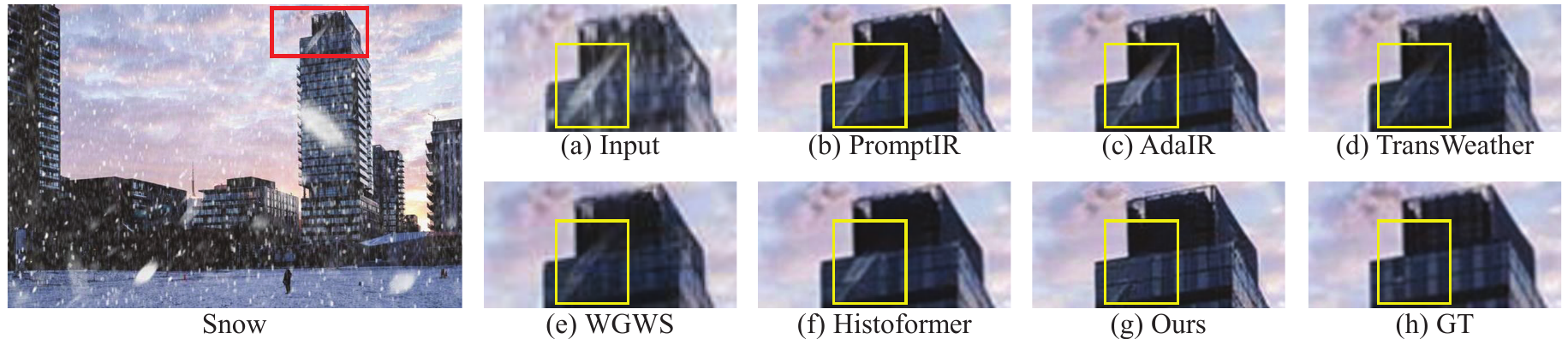}
    \vspace{-7pt}
    \caption{Qualitative results of desnowing Snow100K-L~\cite{liu2018desnownet} dataset.}
    \label{fig:qual_snow}
    \vspace{-14pt}
\end{figure*}

\noindent
\textbf{Qualitative evaluation.}
We compare visualizations with recent methods across three tasks in~\cref{fig:qual_rain},~\ref{fig:qual_drop}, and~\ref{fig:qual_snow}.
Our method restores the original objects more sharply than others and achieves the closest color condition to the clean image. In particular, as shown in~\cref{fig:qual_rain}, the yellow box highlights how our method accurately restores fine objects that are obscured by rain degradation, preventing misclassification as rain artifacts. We can also see that our method fills in the empty areas left after removing degradations rather than others.
The visualization for WeatherStream is provided in the supplementary.

\begin{figure}[tbh]
    \centering
    \includegraphics[width=0.995\linewidth]{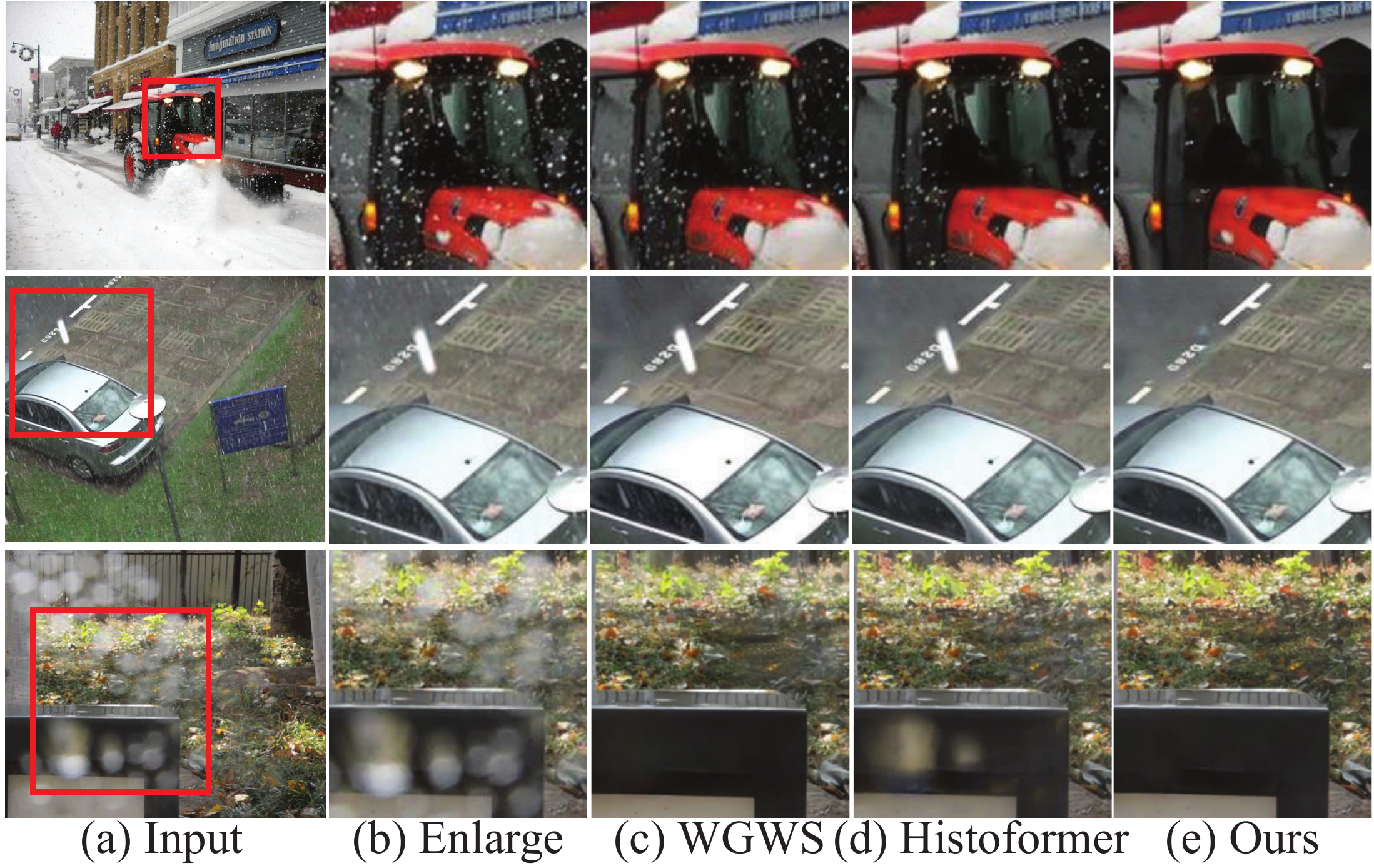}
    \vspace{-20pt}
    \caption{Real-world degradation removal examples. From the first to last rows, the adverse weather type is snow (RealSnow~\cite{zhu2023learning}), rain (NTURain~\cite{chen2018robust}), and, raindrop (RainDS~\cite{quan2021removing}), respectively. Zoom for better view.}
    \vspace{-14pt}
    \label{fig:qual_real}
\end{figure}

\noindent
\textbf{Real-world scenario.}
Additionally, we conduct a visual comparison to evaluate the model trained on the All-Weather dataset under real-world conditions. The results in~\cref{fig:qual_real} show that our method successfully removes real-world degradations better than others.

\subsection{Ablation studies and discussions}
To reveal the relationship between our proposed methods and the degradation removal performance of each weather type, we demonstrate ablation studies with the evaluation in all weather, not specific weather type.

\begin{table}[t!]
    \centering
    \caption{Ablation study for the proposed Spectral-based Decompotision Prompt (SDP).}
    \vspace{-7pt}
    \resizebox{.995\linewidth}{!}{
    \setlength\tabcolsep{5pt}
    \begin{tabular}{C{0.9cm}C{0.8cm}|cc|cc|cc|cc}
    \thickhline
     &  & \multicolumn{2}{c|}{Rain} & \multicolumn{2}{c|}{Snow} & \multicolumn{2}{c|}{Raindrop} & \multicolumn{2}{c}{Avg.} \\
    Sobel & SVD  & PNSR & SSIM & PNSR & SSIM & PNSR & SSIM & PNSR & SSIM\\ \thickhline
    \checkmark  &          & 32.32 & 0.9401 & 32.13 & 0.9259 & \textbf{33.30} & 0.9488 & 32.58 & 0.9383 \\
    &\checkmark            & 32.19 & 0.9400 & 32.08 & 0.9257 & 33.07 & 0.9486 & 32.45 & 0.9381 \\
    \checkmark &\checkmark & \textbf{32.43} &  \textbf{0.9410} & \textbf{32.22} & \textbf{0.9272} & 33.24 & \textbf{0.9489} & \textbf{32.63} & \textbf{0.9390} \\ \thickhline
    \end{tabular}}
    \vspace{-7pt}
    \label{tab:prompt_abl}
\end{table}

\begin{table}[t]
    \centering
    \caption{SGTB Attention configuration. Ch. and Spa. denote channel-wise attention and spatial-wise attention, respectively.}
    \vspace{-7pt}
    \resizebox{.995\linewidth}{!}{
    \setlength\tabcolsep{5pt}
    \begin{tabular}{C{0.8cm}C{0.8cm}|cc|cc|cc|cc}
    \thickhline
    \multicolumn{2}{c|}{}& \multicolumn{2}{c|}{Rain} & \multicolumn{2}{c|}{Snow} & \multicolumn{2}{c|}{Raindrop} & \multicolumn{2}{c}{Avg.} \\
    Ch. & Spa.  & PNSR & SSIM & PNSR & SSIM & PNSR & SSIM & PNSR & SSIM\\ \thickhline
     \checkmark &             & 32.16 & 0.9399 & 31.94 & 0.9239 & \textbf{33.26} & 0.9485 & 32.45 & 0.9374 \\
     \checkmark & \checkmark  & \textbf{32.43} & \textbf{0.9410} & \textbf{32.22} & \textbf{0.9272} & 33.24 &\textbf{0.9489} & \textbf{32.63} & \textbf{0.9390} \\ \thickhline
    \end{tabular}}
    \vspace{-7pt}
    \label{tab:configuration_abl}
\end{table}

\begin{table}[t]
    \centering
    \caption{Group attention configuration. In and Cross denote in-group attention and cross-group attention, respectively.}
        \vspace{-7pt}
    \resizebox{.995\linewidth}{!}{
    \setlength\tabcolsep{5pt}
    \begin{tabular}{C{0.6cm}C{1cm}|cc|cc|cc|cc}
    \thickhline
     \multicolumn{2}{c|}{}& \multicolumn{2}{c|}{Rain} & \multicolumn{2}{c|}{Snow} & \multicolumn{2}{c|}{Raindrop} & \multicolumn{2}{c}{Avg.} \\
    In & Cross  & PNSR & SSIM & PNSR & SSIM & PNSR & SSIM & PNSR & SSIM\\ \thickhline
    \checkmark &            & 32.28 & 0.9403 & 32.00 & 0.9246 & 33.10 & 0.9485 & 32.46 & 0.9378 \\
    \checkmark & \checkmark & \textbf{32.43} & \textbf{0.9410} & \textbf{32.22} & \textbf{0.9272} & \textbf{33.24} & \textbf{0.9489} & \textbf{32.63} & \textbf{0.9390} \\ \thickhline
    \end{tabular}}
    \vspace{-7pt}
    \label{tab:group_abl}
\end{table}

\begin{table}[t]
    \centering
    \caption{
    Ablation Study on Prompt and Transformer Block Choices. Feature-Grouped Attention (FGA) and Spectral-based Decomposition Prompt (SDP) are added progressively.
    }
    \vspace{-7pt}
    \resizebox{.995\linewidth}{!}{
    \setlength\tabcolsep{5pt}
    \begin{tabular}{C{0.8cm}C{0.8cm}|cc|cc|cc|cc}
    \thickhline
    \multicolumn{2}{c|}{} & \multicolumn{2}{c|}{Rain} & \multicolumn{2}{c|}{Snow} & \multicolumn{2}{c|}{Raindrop} & \multicolumn{2}{c}{Avg.} \\
    FGA & SDP & PNSR & SSIM & PNSR & SSIM & PNSR & SSIM & PNSR & SSIM\\ \thickhline
        &    & 31.93 & 0.9386 & 31.92 & 0.9238 & 33.10 & 0.9485 & 32.32 & 0.9369 \\
        & \checkmark & 32.27 & 0.9399 & 32.01 & 0.9249 & 33.19 & 0.9486 & 32.49 & 0.9378 \\ 
    \checkmark &    & 32.18 & 0.9400 & 32.07 & 0.9256 & 33.14 & 0.9486 & 32.46 & 0.9380 \\
    \checkmark & \checkmark & \textbf{32.43} & \textbf{0.9410} & \textbf{32.22} & \textbf{0.9272} & \textbf{33.24} & \textbf{0.9489} & \textbf{32.63} & \textbf{0.9390} \\ \thickhline
    \end{tabular}}
    \vspace{-14pt}
    \label{tab:prompt_choice_abl}
\end{table}

\noindent
\textbf{Design of prompts.} To demonstrate the effectiveness of our proposed SDP, we conduct independent experiments about the components in the SDP. As shown in~\cref{tab:prompt_abl}, using the Sobel operator and SVD together shows the best performance in the degradation type of rain and snow. In raindrop, the SSIM is the highest, and the PSNR is confirmed to be the second best. This experiment demonstrates that the combination of the two decomposition filters can efficiently extract information from the image.

\noindent
\textbf{Spatial Grouping Transformer Block configuration.}
To demonstrate the effectiveness of the proposed SGTB, we conduct experiments by varying its configuration.
In~\cref{tab:configuration_abl}, the proposed configuration is compared with two variants. The first uses SGTB-C for all blocks per stage, where each stage $p$ consists of $L_p$ blocks. The second adopts a hybrid setup, in which half of the blocks per stage are replaced with SGTB-S (\ie, $L_p/2$ SGTB-C and $L_p/2$ SGTB-S).
It is observed that the mixed configuration with SGTB-C and SGTB-S performs better on all metrics.
Notably, this benefit comes without any increase in model capacity.


\noindent
\textbf{Group attention configuration.}
We analyzed about how the formation of attention mechanism is executed within each transformer block in~\cref{tab:group_abl}.
The results showed that incorporating cross-group attention, which enables information exchange between groups, yielded better performance than using in-group attention.

\noindent
\textbf{The synergy between FGA and SDP.}
We conducted an ablation study to evaluate the effects of FGA and SDP, as shown in~\cref{tab:prompt_choice_abl}. Four models were tested for comparison: the baseline model using the standard transformer block, the model using SDP as a prompt, the model using FGA as a transformer block, and the model incorporating both FGA and SDP.
The results demonstrate that the inclusion of SDP significantly improves performance, with the model using SDP outperforming the baseline. This highlights the effectiveness of SDP in extracting degradation information and enhancing restoration. Further improvements are observed with FGA, which enables the model to focus on relevant areas of the scene. For FGA, a simple visual prompt is used to generate a mask, which serves as input. Unlike the baseline model, which lacks a grouping mechanism, this model leverages the mask to group similar areas, leading to better performance, particularly in complex degradation scenarios.
Finally, we conducted an experiment combining both FGA and SDP to assess their synergy. As shown in the last column of~\cref{tab:prompt_choice_abl}, SDP improves restoration by handling degradations, while FGA enhances this process by adding an effective grouping mechanism. The combination of both results in the best overall performance.


\section{Conclusion}
\label{sec:conclusion}
We propose SSGformer, a novel restoration architecture that incorporates Spatial Grouping Transformer Block with Spectral-based Decomposition Prompt. By leveraging the Sobel operator to extract high-frequency edge information and Singular Value Decomposition for low-frequency components, our method effectively captures valuable spectral features from degraded images. With the manipulated features, we generate the grouping-mask. This mask guides the group-wise attention mechanism across both the channel and spatial domains, enabling the model to comprehensively learn degradation-related information across diverse frequency ranges. Through comprehensive experiments, we showcase the effectiveness and advantages of our approach.

\noindent\textbf{Limitations and future works.}
Our method leverages information from classical filters, which makes it effective in scenarios where classical filters are applicable. This design reflects an internal strategy for AiO, relying on intrinsic visual structures to guide restoration. However, in scenes with high background complexity—especially when weather-specific cues are absent—such internal cues may be insufficient. To overcome these limitations, we plan to explore external strategies that incorporate knowledge from large-scale pretrained models, such as LLM/VLMs~\cite{yang2024language}.

\section*{Acknowledgment}
This research was supported by the Challengeable Future Defense Technology Research and Development Program through the Agency For Defense Development(ADD) funded by the Defense Acquisition Program Administration(DAPA) in 2025(No.915102201)

{
    \bibliographystyle{ieeenat_fullname}
    \bibliography{main}
}

\end{document}